\pdfoutput=1
\documentclass[11pt]{article}

\usepackage[final]{acl}

\usepackage{times}
\usepackage{latexsym}

\usepackage[T1]{fontenc}
\usepackage[utf8]{inputenc}

\usepackage{microtype}

\usepackage{inconsolata}

\usepackage{graphicx}
\usepackage{todonotes}
\usepackage{xcolor}         % colors
\usepackage{svg}
\usepackage{booktabs}
\usepackage{amsmath, amssymb, amsfonts}
\usepackage{subcaption}
\usepackage{multirow}
\title{Explainable Chain-of-Thought Reasoning: An Empirical Analysis on State-Aware Reasoning Dynamics}

\author{
  \textbf{Sheldon Yu}$^{1}$\thanks{\;Equal contribution.} \quad
  \textbf{Yuxin Xiong}$^{1}$\footnotemark[1] \quad
  \textbf{Junda Wu}$^{1}$ \quad
  \textbf{Xintong Li}$^{1}$ \quad
  \textbf{Tong Yu}$^{2}$ \\
  \textbf{Xiang Chen}$^{2}$ \quad
  \textbf{Ritwik Sinha}$^{2}$ \quad
  \textbf{Jingbo Shang}$^{1}$ \quad
  \textbf{Julian McAuley}$^{1}$ \\
  $^{1}$University of California San Diego \quad
  $^{2}$Adobe Research \\
  \texttt{\{ziy040, y7xiong, juw069, xil240, jshang, jmcauley\}@ucsd.edu} \\
  \texttt{\{tyu, xiangche, risinha\}@adobe.com}
}

\begin{document}
% \renewcommand{\headrulewidth}{0pt}

% Proceedings footer line (watermark-style)
% \AddToShipoutPicture*{%
%   \setlength{\unitlength}{1mm}%
%   \footnotesize%
%   \put(0,13){\parbox[t]{\paperwidth}{\centering
%     \emph{\ProceedingsTitle}, pages \Pages \\
%     \ConferenceDatesAndPlace, \ConferenceYear\ \textcopyright\ \ConferenceYear\ \Publisher
%   }}%
% }

% Include your already-built paper PDF here:
% \includepdf[
%   pagecommand={\thispagestyle{plain}},
%   pages=-
% ]{Explainable_CoT (3).pdf}
\maketitle

\begin{abstract}
% Recent advances in chain-of-thought (CoT) prompting have demonstrated the ability of large language models (LLMs) to perform multi-step reasoning. While prior work focuses on improving CoT generation quality or attributing token-level importance, we propose a novel framework to structurally analyze the latent dynamics of CoT trajectories for interpretability. Our method segments generated CoT into discrete reasoning steps, abstracts each step into a spectral embedding based on the eigenvalues of token-level Gram matrices, and clusters these embeddings into semantically meaningful latent states. We model the global evolution of reasoning as a first-order Markov chain over latent clusters, yielding interpretable transition structures. Through t-SNE visualizations and Monte Carlo rollouts, we uncover consistent trajectories across tasks and models, supporting the hypothesis that LLM reasoning follows globally coherent yet abstract paths. 

Recent advances in chain-of-thought (CoT) prompting have enabled large language models (LLMs) to perform multi-step reasoning. 
However, the explainability of such reasoning remains limited,
with prior work primarily focusing on local token-level attribution, 
such that the high-level semantic roles of reasoning steps and their transitions remain underexplored. 
In this paper, we introduce a state-aware transition framework that abstracts CoT trajectories into structured latent dynamics. 
Specifically, to capture the evolving semantics of CoT reasoning,
each reasoning step is represented via spectral analysis of token-level embeddings and clustered into semantically coherent latent states. 
To characterize the global structure of reasoning, we model their progression as a Markov chain, 
yielding a structured and interpretable view of the reasoning process. 
This abstraction supports a range of analyses, including semantic role identification, temporal pattern visualization, and consistency evaluation.

\end{abstract}

\section{Introduction}

Chain-of-thought (CoT) prompting has become a central technique for eliciting multi-step reasoning in large language models (LLMs) \cite{wei2022chain}. 
By encouraging models to decompose problems into intermediate steps, CoT improves performance on tasks such as arithmetic, logical deduction, and multi-hop question answering.
Understanding CoT outputs is, therefore, critical for both evaluation and user comprehension. Recent large-scale studies have shown that even modest increases in textual complexity can hinder human comprehension and increase cognitive load, highlighting the need for more interpretable LLM outputs~\cite{guidroz2025llm,xiao2025streaming,wasi2024cogergllm,wu2025ctrls,wu2024ocean,wu2024decot}. 
Understanding CoT is even more challenging in this context, given its length, multi-step structure, and abstract semantic progression.
However, the explainability of CoT remains limited. 
Prior work often focuses on token-level attribution\cite{hou2023mechanistic} or heuristic correctness measures\cite{gan2025rethink}.
Focusing on local token-level attribution leaves the high-level semantic roles of reasoning steps and their transitions underexplored. 
Understanding CoT reasoning benefits from a shift from local analysis to a structured, global perspective.

%We argue that understanding CoT requires a shift from local analysis to a structured, global perspective.
% Rather than viewing a CoT as a flat sequence of text, we model it as a trajectory through a sequence of latent semantic states, where each state captures the evolving function of a reasoning step. This abstraction enables higher-level questions: How do reasoning strategies differ across models or tasks? Are there consistent structural patterns in CoT trajectories? 
To address this, we propose a \textbf{state-aware transition framework} for CoT reasoning. We segment generated CoTs into discrete steps, embed each step via spectral analysis of token-level representations, and cluster them into semantically coherent latent states. Transitions between these states are modeled as a first-order Markov chain, yielding a structured view of reasoning dynamics. This approach goes beyond surface-level generation by uncovering latent structure in CoT trajectories, offering a non-trivial solution to the challenge of interpreting abstract, multi-step reasoning without access to ground-truth annotations or explicit reasoning labels.
Designed as an explainability scaffold, this framework supports diverse analyses, including semantic role identification, temporal transition pattern discovery, and trajectory-level consistency evaluation.
We summarize our contributions as follows:

% \noindent\textbf{Our contributions are as follows:}
% \vspace{0.5em}
% \begin{itemize}
%   \item We introduce a novel abstraction of CoT reasoning as transitions over latent semantic states, enabling structured analysis of reasoning behaviors.
%   \item We develop a spectral embedding pipeline that maps reasoning steps into interpretable state trajectories via clustering and Markov modeling.
%   \item We demonstrate how this framework supports multiple interpretability applications, such as role discovery, structure visualization, and consistency diagnostics.
%   \item Empirical results across tasks and models reveal consistent reasoning patterns and latent structure beyond surface-level generation.
% \end{itemize}

\begin{itemize}

  %\item We propose a novel abstraction of CoT reasoning as structured transitions over latent semantic states, enabling a global and interpretable view of multi-step reasoning.
  %\item We develop a spectral embedding pipeline based on eigenvalue descriptors, followed by unsupervised clustering and Markov modeling.
      \item We propose a framework that abstracts CoT reasoning as structured transitions over latent semantic states, modeled via a Markov Chain.
    
      \item We further visualize multi-step CoT reasoning for explainable understanding of reasoning dynamics beyond token-level attribution.
      
  %\item We demonstrate that this abstraction supports interpretability use cases such as semantic role discovery, structure visualization, and trajectory-level diagnostics.
  
  %\item Empirical results across tasks and models reveal consistent latent dynamics beyond token-level reasoning.
    
    \item  We enable explainabilities for CoT, including reasoning step abstraction, semantics, and state-aware transition.
      
    \item Empirical results across multiple tasks and models reveal consistent and recurring latent transition patterns, suggesting that LLM reasoning exhibits structured dynamics beyond surface-level token sequences.
  
\end{itemize}

 % \tong{please make sure the claimed itemized contributions can be grounded to the contributions mentioned in the abstract, method, and experiments (i.e., use the same terminologies for the key concepts related to the paper contributions)}

\begin{figure*}[t]
  \centering
  \includegraphics[width=\textwidth, clip, trim=0 170 0 150]{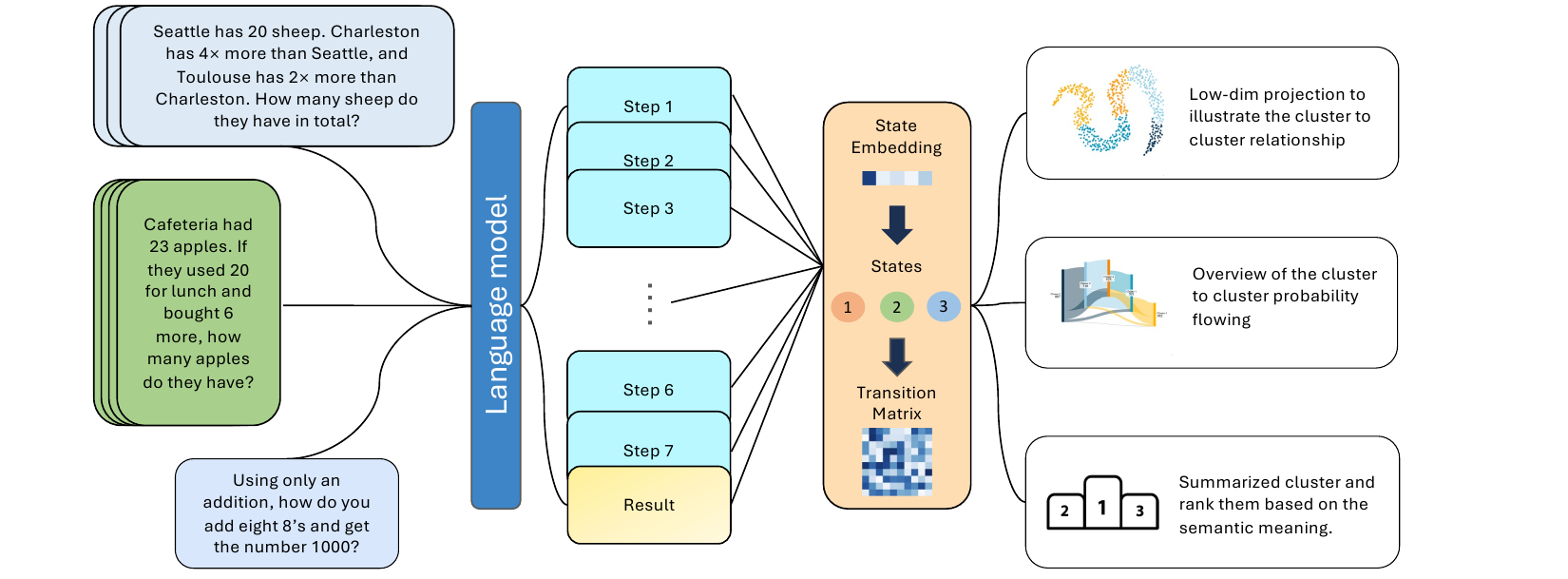}
  \caption{An overview of our CoT abstraction and simulation framework.}
  \label{fig:overview}
  \vspace{-1em}
\end{figure*}

% \improve{add one overview fig}
\section{Related Work}

% \item Multi-step Problem Solving Through a Verifier: An Empirical Analysis on Model-induced Process Supervision.

% \item On the Empirical Complexity of Reasoning and Planning in LLMs.

% \item An Empirical Analysis on Large Language Models in Debate Evaluation.
\noindent\textbf{Chain-of-Thought (CoT) Reasoning in LLMs.}  
Chain-of-Thought (CoT) prompting has emerged as an effective strategy to elicit multi-step reasoning in large language models (LLMs)~\cite{wei2022chain}. Subsequent work explored improved prompting techniques~\cite{kojima2023large}, automatic CoT generation~\cite{zhang2022auto}, and evaluation of reasoning quality~\cite{wang2023does}. While most prior work focuses on generating or enhancing CoT outputs, we aim to abstract and model the latent structure of CoTs for interpretability. Coconut~\cite{hao2024training} introduces a latent reasoning paradigm that operates in continuous space rather than natural language, enabling breadth-first exploration of reasoning paths and highlighting the limitations of token-level CoT representations. 
This further motivates our structural approach to modeling CoT dynamics.

\noindent\textbf{Explainability in Language Models.}  
Explainability methods for LLMs often focus on token-level attribution, such as attention rollout~\cite{chuang2024lookback,wu2021deconfounded} or integrated gradients~\cite{zhao2024towards}. 
In addition, in-context explanability~\citep{liu2023towards,wei2023larger,liu2024interact,wu2022context} focuses on understanding how to best prompt LLMs with demonstrations.
Recent work has also questioned the faithfulness of CoT explanations~\cite{turpin2023language,atanasova2023faithfulness}. 
From the perspective of LLM's latent space, the gradient-based method~\cite{wu2023analyzing} explains more fine-grained CoT behaviors.
In contrast, we propose a structural abstraction framework that operates at the step level, capturing global reasoning dynamics rather than local token importance.
\section{Explainable Modeling of Chain-of-Thought Reasoning}

\noindent\textbf{Step Embedding and Abstraction}
To capture the evolving semantics of each reasoning step, we segment the CoT output \( y \) (elicited from an instruction-tuned LLM given input \( x \)) into a sequence of steps \( c = (c_1, \dots, c_T) \). For each step \( c_t \), 
we extract token embeddings \( X_t \in \mathbb{R}^{n_t \times d} \) and compute the local Gram matrix \( \tilde{G}_t = X_t^\top X_t \).
We define the spectral embedding \( E_t \in \mathbb{R}^{k_{\text{eig}}} \) as the vector of top-\( k_{\text{eig}} \) eigenvalues:
\[
E_t = (\lambda_1, \dots, \lambda_{k_{\text{eig}}}), \quad \lambda_1 \geq \cdots \geq \lambda_{k_{\text{eig}}}.
\]
To incorporate contextual accumulation, we recursively update the accumulated Gram matrix across steps:
\[
G_t = G_{t-1} + \tilde{G}_t, \quad G_1 = \tilde{G}_1.
\]
This yields a trajectory of spectral embeddings \( (E_1, \dots, E_T) \), which we abstract using \(k\)-means with \(k_{\text{clu}}\) clusters to assign each step to a latent state \( s_t \in \{1, \dots, k_{\text{clu}}\} \).
The resulting state sequence enables structural abstraction of CoT and supports distribution-level explainability.

\noindent\textbf{Semantics of Reasoning State}
To assess whether latent clusters correspond to meaningful functional roles, we collect all reasoning steps assigned to each cluster and manually summarize their semantics.
To further ground these semantics, we compute the average position of steps assigned to each cluster within the reasoning trajectory. Specifically, for cluster \( c \), the average step index is given by:
\[
\bar{t}_c = \frac{1}{|S_c|} \sum_{(i,t) \in S_c} t,
\]
where \( S_c \) denotes the set of steps assigned to cluster \( c \), and \( t \) is the position of the step in its trajectory. 
% This quantifies whether certain clusters tend to appear earlier or later. 

\noindent\textbf{Modeling Transitions via Markov Chains}
To capture the global structure of CoTs, 
we model transitions between clusters as a Markov chain. 
Given the step-wise state sequence \( \mathbf{s} = (s_0, \dots, s_T) \),
we estimate a transition matrix \( P \in \mathbb{R}^{k \times k} \), where each entry denotes:
\[
P_{i,j} = \mathbb{P}(s_{t+1} = j \mid s_t = i) = \frac{C_{i,j}}{\sum_{j'} C_{i,j'}},
\]
with \( C_{i,j} \) as the number of observed transitions from state \( i \) to \( j \). 
This latent transition structure reveals sequential patterns in CoT and serves as the basis for downstream diagnostics and simulation.

\section{Analysis of Reasoning Dynamics}
% \tong{The section title might be improved to better highlight the contributions.}
\noindent\textbf{Datasets}
% \tong{It would be helpful to add a table when presenting the experiment results.}
To comprehensively assess the structural dynamics of chain-of-thought (CoT) reasoning, we consider datasets from three distinct categories: 
\textbf{mathematical} GSM8k~\cite{cobbe2021gsm8k}, MATH~\cite{hendrycks2021measuring}, focusing on symbolic and numerical reasoning; \textbf{knowledge-based} (HotpotQA~\cite{yang2018hotpot}, MusiQUe~\cite{trivedi2022musique}), involving multi-hop factual inference over text; and \textbf{commonsense} (CSQA~\cite{talmor2019common}, SocialIQa~\cite{sap2019social}), targeting intuitive reasoning about everyday and social scenarios. This diverse selection enables us to evaluate whether latent reasoning structures generalize across quantitative, textual, and intuitive domains.

\noindent\textbf{Implementation Details}
We analyze CoT reasoning using three instruction-tuned language models: Gemma 2B~\cite{gemma}, LLaMA 3.2B~\cite{liu2025spin}, and Qwen2.5 7B~\cite{qwen2025qwen}. For each question, we generate a CoT response using a fixed prompt and segment it into reasoning steps via explicit textual markers. We extract token-level hidden states, compute cumulative Gram matrices, and obtain spectral embeddings by taking the top-64 eigenvalues. These embeddings are clustered into \( k{=}5 \) latent states via \( k \)-means to estimate a first-order transition matrix \( P \). Monte Carlo rollouts from \( P \) yield synthetic reasoning trajectories.

% \subsection{Reasoning Step Explainability}

% To examine whether the latent clusters correspond to semantically meaningful reasoning behaviors, we analyze the content of each cluster by aggregating step texts from GSM8k and manually summarizing their functional roles. As shown in Table~\ref{tab:cluster-summary-1}, the clusters align with interpretable categories such as problem setup, intermediate computation, and final aggregation.

% \subsection{Distributional Explainability of Reasoning Steps}

% We first examine whether reasoning steps form structurally meaningful clusters in the latent embedding space. Using spectral embeddings $\{E_t\}$ extracted from GSM8K, we visualize the step distribution with t-SNE and color points by their cluster assignment. As shown in Figure~\ref{fig:tsne}, the clusters exhibit clear separation with minimal overlap, suggesting that the latent space captures distinct structural modes of reasoning. Notably, the same clustering patterns emerge across tasks such as SocialIQA and MuSiQue, indicating robustness of the abstraction.

\subsection{Explainability of Reasoning Abstraction}

\begin{figure}[t]

    \begin{subfigure}[t]{\columnwidth}
    \centering
    \includegraphics[trim=0 0 0 0,width=0.8\columnwidth]{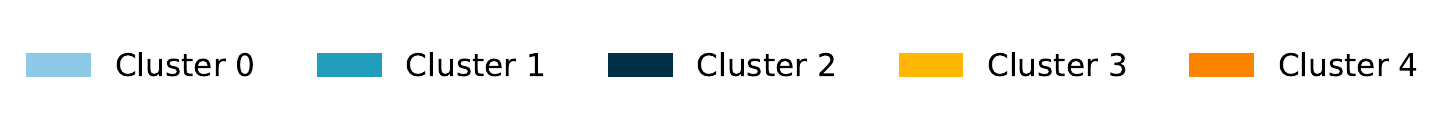}
  \end{subfigure}
  \centering
  % first row
  \begin{subfigure}[b]{0.5\columnwidth}
    \includegraphics[width=\linewidth]{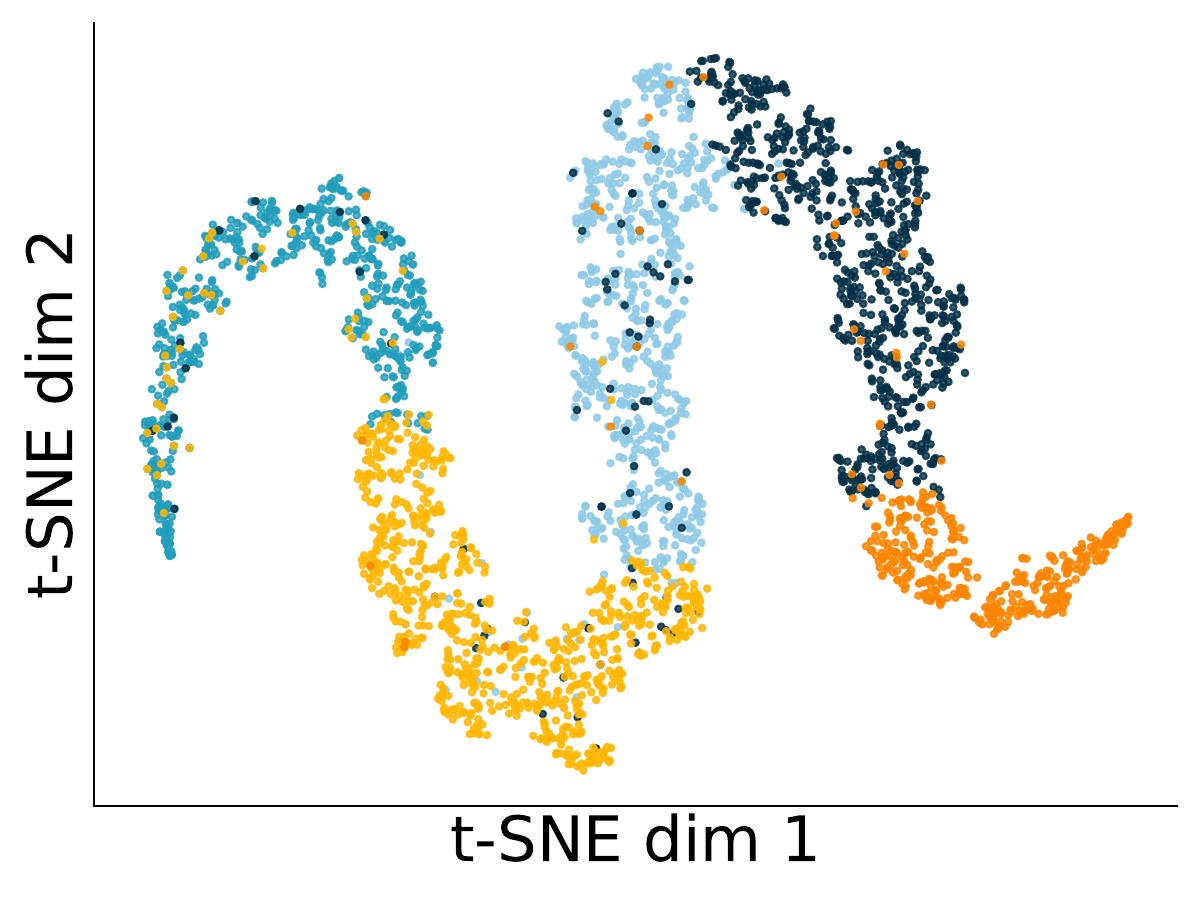}
    \caption{Llama on GSM8K}
    \label{fig:tl}
  \end{subfigure}\hfill
  \begin{subfigure}[b]{0.5\columnwidth}
    \includegraphics[width=\linewidth]{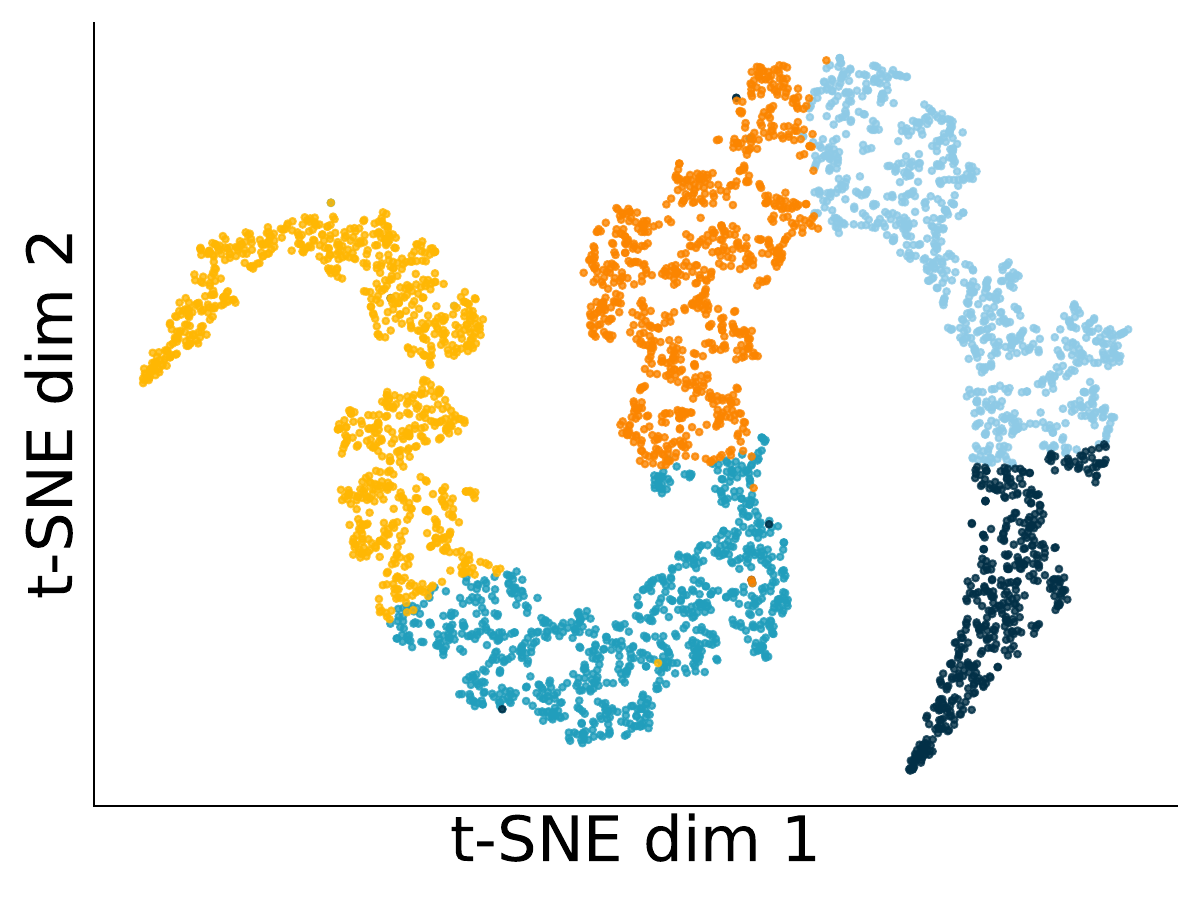}
    \caption{Llama on SocialIQA}
    \label{fig:tr}
  \end{subfigure}
  % second row
  \begin{subfigure}[b]{0.5\columnwidth}
    \includegraphics[width=\linewidth]{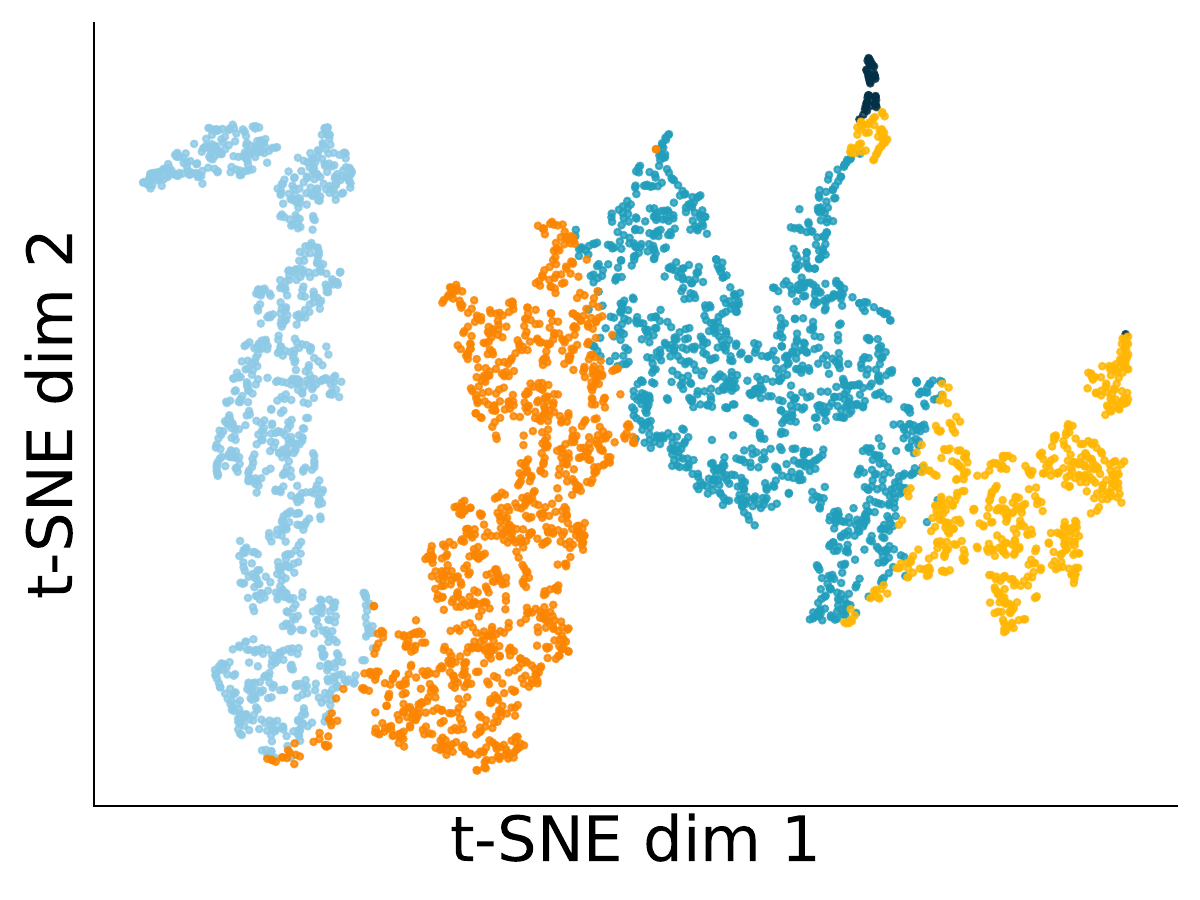}
    \caption{Llama on on Math}
    \label{fig:bl}
  \end{subfigure}\hfill
  \begin{subfigure}[b]{0.5\columnwidth}
    \includegraphics[width=\linewidth]{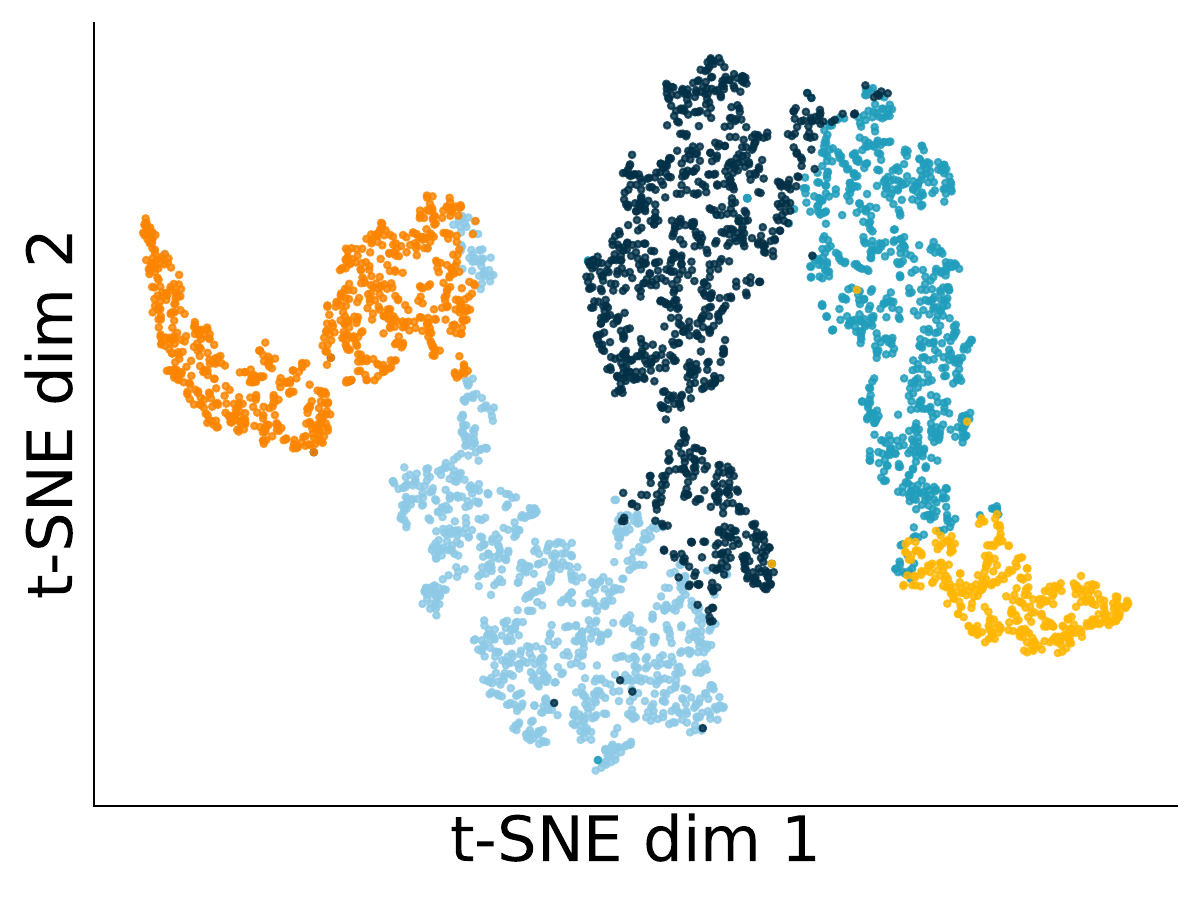}
    \caption{Llama on MusiQUe}
    \label{fig:br}
  \end{subfigure}

  \caption{t-SNE projection of Chain-of-Thought step embeddings from Llama-3B across four datasets.}
  \label{fig:tsne}
  \vspace{-1em}
\end{figure}

We examine whether reasoning steps exhibit structurally coherent organization in the latent embedding space.
For each step, we compute its spectral embedding and apply t-SNE for visualization. 
Figure~\ref{fig:tsne} shows the resulting projections across four datasets, GSM8K, SocialIQA, Math, and MuSiQue, using LLaMA-3B.
Coloring each point by its cluster assignment reveals clear separation with minimal overlap, suggesting that the latent representations capture distinct structural modes of reasoning.
Notably, the same clustering patterns consistently emerge across diverse domains, indicating that the abstraction is robust and generalizable beyond a single task. 
These visualizations provide explainability by revealing how reasoning steps naturally fall into functionally distinct groups, even without explicit supervision.

% \subsection{Reasoning Step Explainability \tong{The subsection title is the same as the previous subsection title}}

\begin{table*}[t]
\centering
\small
\begin{tabular}{c c c c}
\toprule
\textbf{Cluster ID} & \textbf{Label} & \textbf{Description} &\textbf{Rank in CoT} \\
\midrule
0 & Scenario Description & Summarize the scenario’s characters, actions, setting. & 4 \\
1 & Problem Framing & Extract key facts and define the question and options. & 2 \\
2 & Detailed Option Evaluation & Integrate reasoning steps into a clear conclusion. & 5\\
3 & Option Analysis & Laying out settings before evaluating choices. & 1\\
4 & Answer Synthesis & Systematically align each option with the context.		 & 3 \\
\bottomrule
\end{tabular}
\caption{Interpretation of unsupervised cluster assignments on LLaMA-generated reasoning steps from SocialIQA.}
\label{tab:cluster-summary}
  \vspace{-1em}
\end{table*}

% To further illustrate the semantic separation, we visualize the spectral embeddings $\{E_t\}$ using t-SNE and color the points by their cluster assignments. The resulting plot~\ref{fig:tsne} shows distinct clusters with minimal overlap, suggesting that the learned step embeddings capture structurally different modes of reasoning. This supports the hypothesis that CoT reasoning follows discrete semantic stages even without explicit supervision.
\subsection{Explainability of Reasoning Semantics}

% To investigate whether latent clusters align with interpretable reasoning behavior, we aggregate the texts of all steps within each cluster and manually summarize their functional roles. Table~\ref{tab:cluster-summary} presents the annotated labels and descriptions for clusters derived from LLaMA-generated CoT steps on SocialIQA. The clusters correspond to intuitive categories such as scenario description, problem framing, option evaluation, and answer synthesis. These functional interpretations suggest that unsupervised latent clustering can reveal the semantic organization of reasoning steps, even without access to explicit supervision.

% % add the step index

% We further validate the semantic interpretation by examining the average step index of each cluster across models. As shown in Table~\ref{tab:step-index}, the cluster IDs exhibit a strong alignment with the natural temporal order of reasoning: clusters with lower IDs (e.g., C0, C2) dominate early steps, while higher-indexed clusters (e.g., C4) tend to appear toward the end. This ordering effect is particularly consistent for LLaMA, matching our manual summary in Table~\ref{tab:cluster-summary}, where early clusters correspond to scenario description and problem framing, and later clusters correspond to answer synthesis. 

To investigate whether latent clusters correspond to meaningful reasoning behaviors,
we aggregate the texts of all steps within each cluster and manually summarize their functional roles.
As shown in Table~\ref{tab:cluster-summary}, clusters derived from LLaMA-generated CoT on SocialIQA align with intuitive categories such as scenario description, problem framing, option evaluation, and answer synthesis. 
These roles were annotated based on step content and ranked by their typical position in the CoT trajectory.

We further validate these interpretations by computing the average step index for each cluster. Table~\ref{tab:sim_real_full} reports the average step index of each cluster based on real CoT trajectories across models and datasets. For LLaMA, we observe strong consistency: clusters with lower average indices (e.g., C3, C1) correspond to early-stage functions like option analysis and problem framing, while those with higher indices (e.g., C2, C4) align with final-stage synthesis. This ordering effect mirrors our manual rank assignment in Table~\ref{tab:cluster-summary}, reinforcing the interpretability of the latent abstraction.

% \begin{itemize}
%     \item @Sheldon: think of a visualization to show: GPT summarize each clustering into some action abstraction.
%     \item visualize clustering of thoughts via T-SNE
%     \item @Sheldon: if spectrum clustering has it's own visualization method
% \end{itemize}

\subsection{Explainability of Reasoning Transition}

\begin{figure}[t]
  \centering
  \includegraphics[trim=0 0bp 0 0bp, width=0.9\columnwidth]{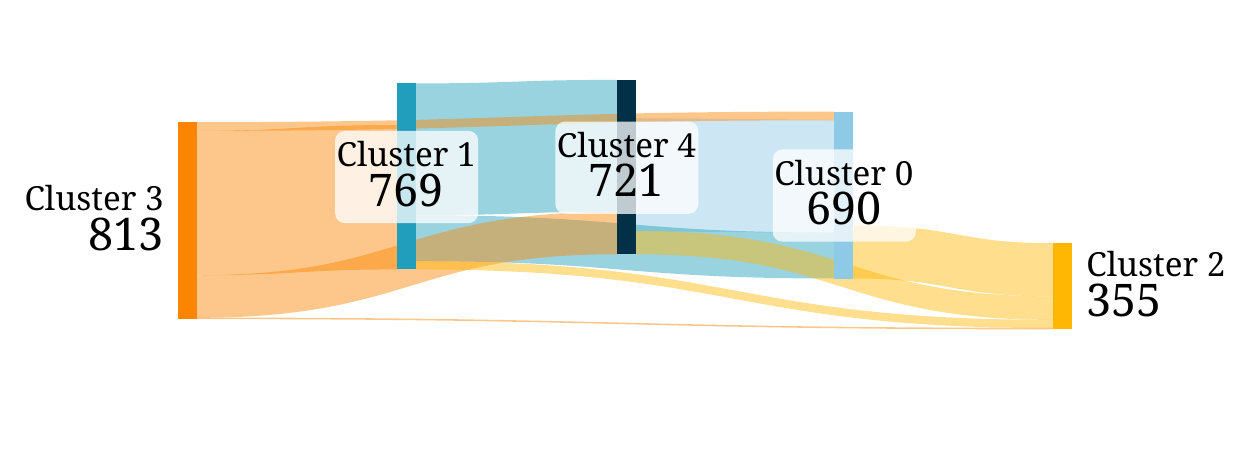}         
  \includegraphics[trim=0 0bp 0 50bp,width=0.9\columnwidth]{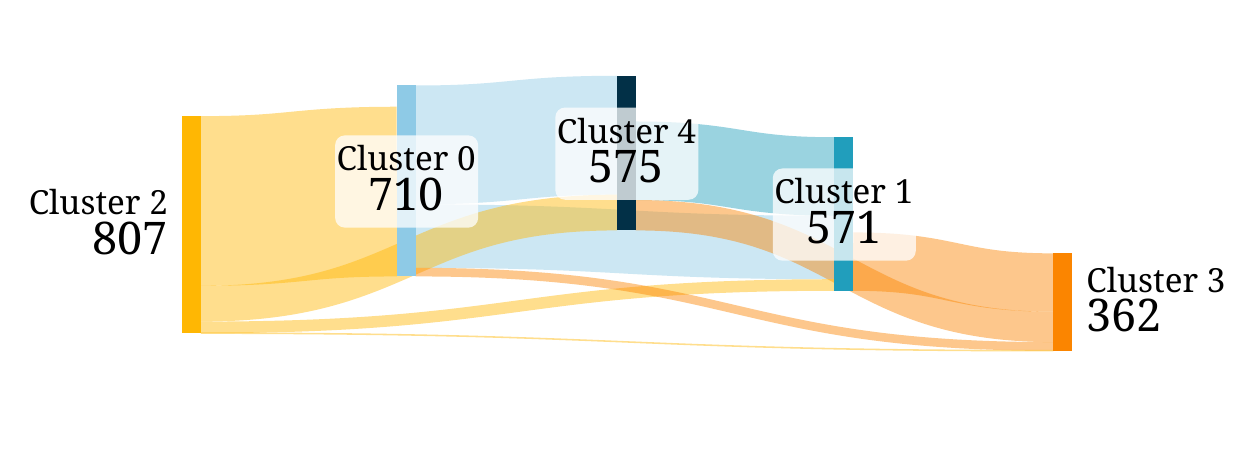}
  \includegraphics[trim=0 0bp 0 50bp,width=0.9\columnwidth]{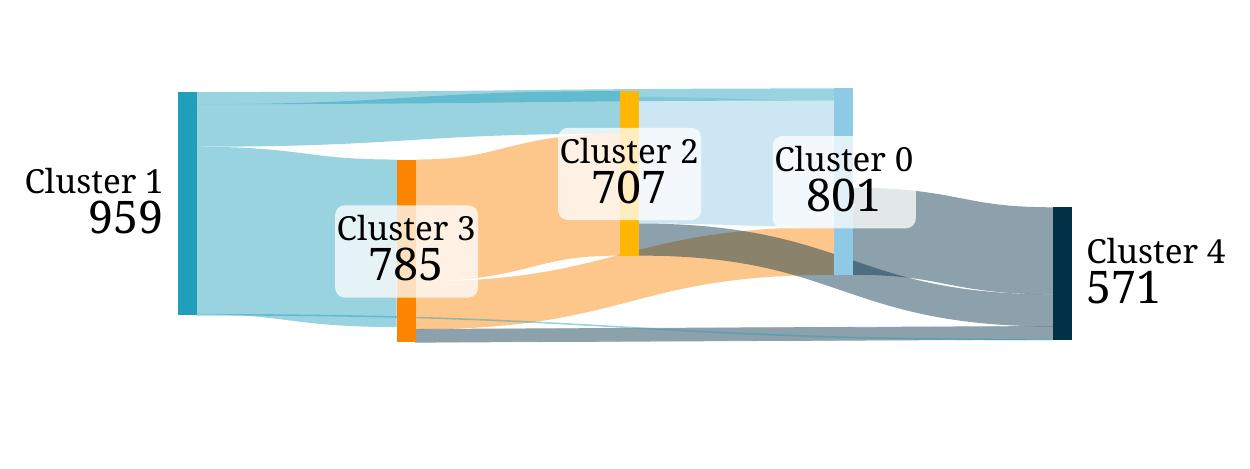}
  \caption{Transition diagrams for the SocialIQA dataset. Each diagram visualizes one model: Llama-3 3B (top), Gemma-2 2B (middle) and Qwen-7B (bottom)}
  \label{fig:stacked}
  \vspace{-1em}
\end{figure}

% Given the clustered state sequences $\{z_t\}$, we estimate a first-order Markov transition matrix $P \in \mathbb{R}^{k \times k}$ to characterize the dynamics of reasoning flows. We visualize $P$ as a heatmap (Fig.~\ref{fig:heatmap}, revealing asymmetric transition patterns—for instance, certain clusters (e.g., setup) tend to initiate reasoning, while others (e.g., aggregation) dominate later steps.

% To quantify the temporal roles of each cluster, we compute the expected step index at which each cluster appears in real CoT trajectories. This highlights clusters that typically occur at the beginning, middle, or end of reasoning, and offers a coarse alignment between latent states and reasoning phases.

% We further visualize the most frequent transition paths as flow charts (e.g., Sankey diagrams), which provide an interpretable summary of global reasoning dynamics across the dataset. These diagrams highlight dominant CoT pathways such as \texttt{setup}~$\rightarrow$~\texttt{computation}~$\rightarrow$~\texttt{aggregation}, and make it easier to compare reasoning patterns across models or tasks.

\begin{figure}[t]
  \centering
  % first row
  \begin{subfigure}[b]{0.5\columnwidth}
    \includegraphics[width=\linewidth]{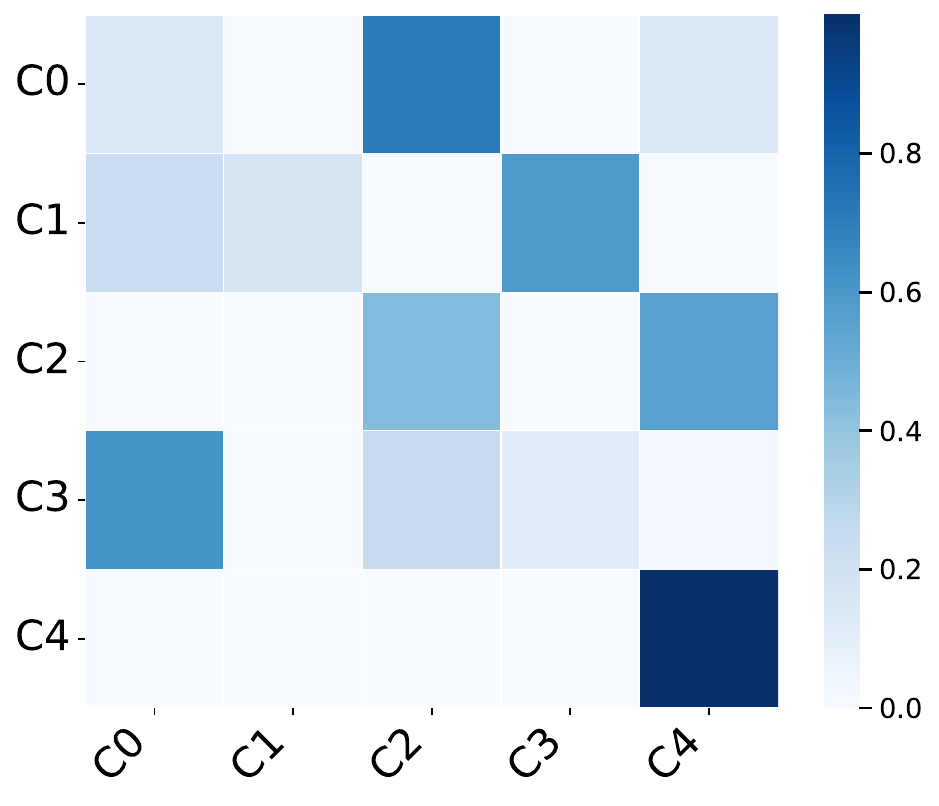}
    \caption{Llama on GSM8K}
    \label{fig:tl}
  \end{subfigure}\hfill
  \begin{subfigure}[b]{0.5\columnwidth}
    \includegraphics[width=\linewidth]{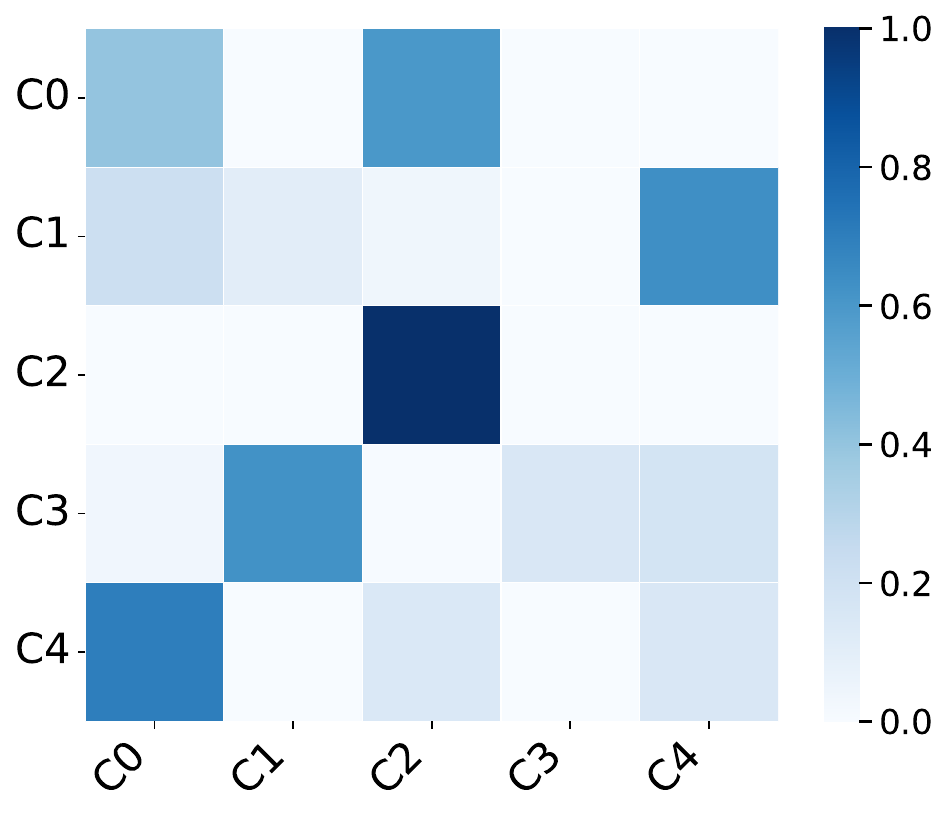}
    \caption{Llama on SocialIQA}
    \label{fig:tr}
  \end{subfigure}

  \vspace{1ex} % small gap between rows

  % second row
  \begin{subfigure}[b]{0.5\columnwidth}
    \includegraphics[width=\linewidth]{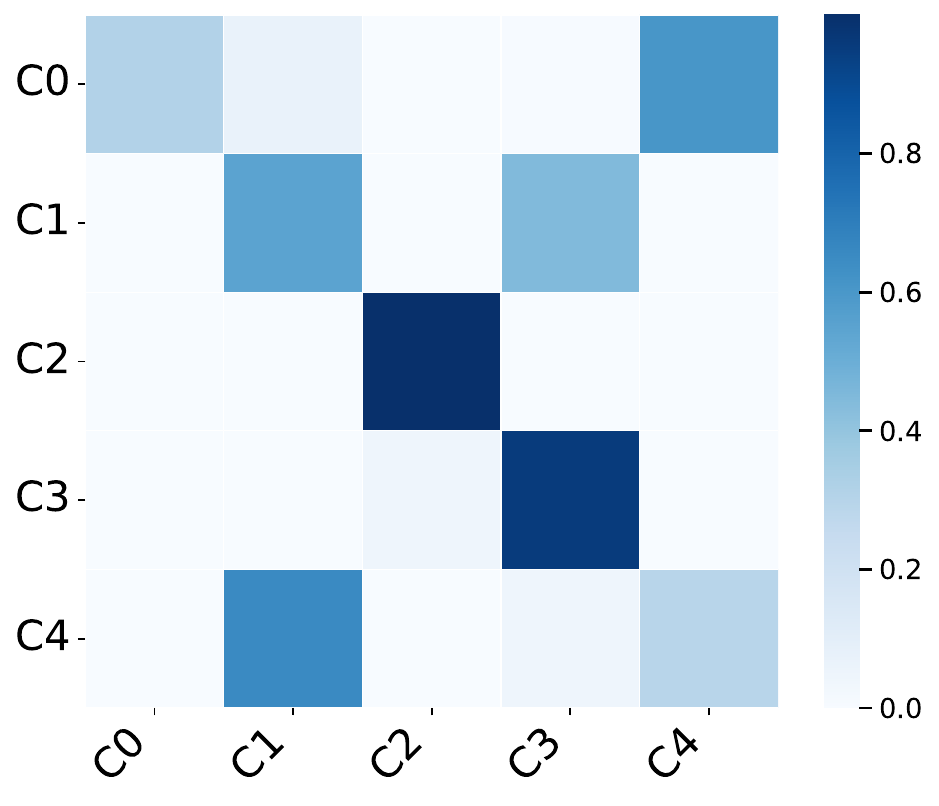}
    \caption{Llama on on Math}
    \label{fig:bl}
  \end{subfigure}\hfill
  \begin{subfigure}[b]{0.5\columnwidth}
    \includegraphics[width=\linewidth]{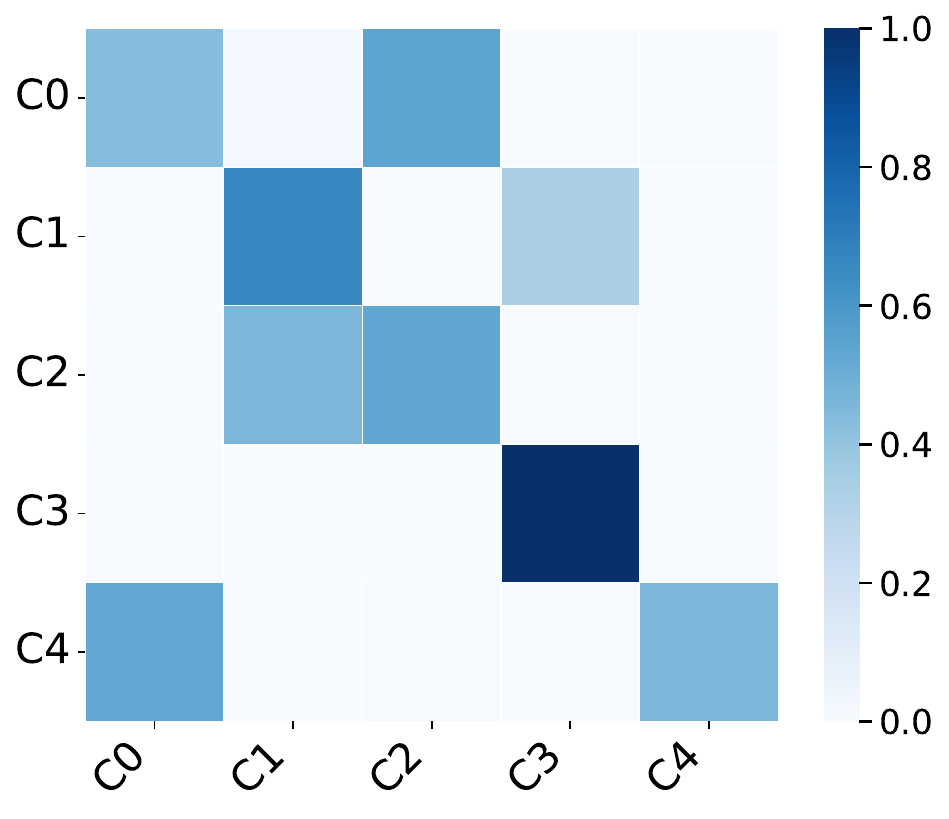}
    \caption{Llama on on MusiQue}
    \label{fig:br}
  \end{subfigure}

  \caption{Cluster transition probability heatmaps for Llama-3B across four benchmarks.}
  \label{fig:heatmap}
  \vspace{-1em}
\end{figure}

To capture the structural dynamics of CoT reasoning, 
we model the transitions between latent states as a Markov chain.
Given the clustered state sequence \( \{z_t\} \), we estimate a transition matrix \( P \in \mathbb{R}^{k \times k} \), 
where each entry \( P_{i,j} \) reflects the empirical probability of transitioning from state \( i \) to \( j \). 
We visualize \( P \) as a heatmap (Figure~\ref{fig:heatmap}), 
which reveals structured and asymmetric transition patterns.
For instance, certain clusters (e.g., setup-related) predominantly initiate trajectories, 
while others (e.g., synthesis) absorb transitions at the end, suggesting coherent behaviors across tasks.

To further highlight dominant reasoning trajectories, we render the most frequent transitions using Sankey diagrams (Figure~\ref{fig:stacked}) for models like LLaMA, Gemma, and Qwen on SocialIQA. These visualizations show that most trajectories follow consistent paths such as \texttt{scenario description}~$\rightarrow$~\texttt{option evaluation}~$\rightarrow$~\texttt{answer synthesis}, which aligns well with our semantic interpretation in Table~\ref{tab:cluster-summary}. This correspondence provides evidence that the learned transition structure reflects meaningful reasoning phases, rather than arbitrary or noisy transitions.

We also compute the expected step index at which each cluster appears to ground these roles temporally. As shown in Table~\ref{tab:sim_real_full}, the observed ordering of clusters is highly correlated with real reasoning positions, particularly for LLaMA. This confirms that the latent transition model captures natural CoT reasoning.

% \improve{We need the expected step index at which each cluster appears in real CoT trajectories and the most frequent transition paths as flow charts.}

% \input{latex/4_validation}
\section{Conclusion}

%We conclude that ....
We proposed a state-aware transition framework that abstracts CoT trajectories into structured latent dynamics, offering a global perspective on multi-step reasoning in LLMs. Each reasoning step is embedded via spectral analysis and clustered into semantically coherent states, with their transitions modeled as a first-order Markov chain. This approach moves beyond token-level attribution by uncovering consistent latent structures across models and tasks. It enables explainability applications such as semantic role identification and temporal pattern visualization.Empirical results demonstrate that LLMs exhibit structured reasoning patterns, pointing to underlying strategies beyond surface-level token sequences.
\section{Limitation}
Our framework assumes access to internal representations of open-source language models. 
This assumption is common in many existing interpretability and reasoning analysis works~\cite{bharadwaj2024understanding, tang2025unlocking,orlicki2025beyond}.
While our framework provides interpretable abstractions of CoT reasoning, it focuses primarily on intrinsic structural analysis.
% As with prior work on CoT explanations~\cite{turpin2023language, wang2023towards, zhang2022auto}, we do not directly evaluate the impact of these abstractions on downstream task performance, which remains an interesting direction for future research.

% \tong{Please double check the above limitations. If mentioning downsteam tasks is risky, please remove the second limitation.}

%Our framework models CoT reasoning as latent state transitions, enabling multiple forms of analysis. However, it shares several limitations with existing work.

%First, the spectral embeddings are based on token-level representations from frozen LLMs, which may not fully capture semantic equivalence between steps. This limitation is common across many unsupervised CoT analysis methods.

%Second, the clustering assumes discrete semantic states, which may oversimplify the nuanced nature of reasoning. Prior works adopting step-level abstraction face similar trade-offs between interpretability and granularity.

%Lastly, while we use Monte Carlo rollouts to probe transition dynamics, this remains an intrinsic evaluation. As with prior CoT reasoning studies, downstream validation is left for future work.

\section{Acknowledgment}
This work is partially supported by NSF IIS-2432486

\bibliography{main}

@article{orlicki2025beyond,
  title={Beyond Words: A Latent Memory Approach to Internal Reasoning in LLMs},
  author={Orlicki, Jos{\'e} I},
  journal={arXiv preprint arXiv:2502.21030},
  year={2025}
}

@article{bharadwaj2024understanding,
  title={Understanding Hidden Computations in Chain-of-Thought Reasoning},
  author={Bharadwaj, Aryasomayajula Ram},
  journal={arXiv preprint arXiv:2412.04537},
  year={2024}
}

@article{tang2025unlocking,
  title={Unlocking General Long Chain-of-Thought Reasoning Capabilities of Large Language Models via Representation Engineering},
  author={Tang, Xinyu and Wang, Xiaolei and Lv, Zhihao and Min, Yingqian and Zhao, Wayne Xin and Hu, Binbin and Liu, Ziqi and Zhang, Zhiqiang},
  journal={arXiv preprint arXiv:2503.11314},
  year={2025}
}

@article{turpin2023language,
  title={Language models don't always say what they think: Unfaithful explanations in chain-of-thought prompting},
  author={Turpin, Miles and Michael, Julian and Perez, Ethan and Bowman, Samuel},
  journal={Advances in Neural Information Processing Systems},
  volume={36},
  pages={74952--74965},
  year={2023}
}

@article{xiao2025streaming,
  title={Streaming, Fast and Slow: Cognitive Load-Aware Streaming for Efficient LLM Serving},
  author={Xiao, Chang and Yang, Brenda},
  journal={arXiv preprint arXiv:2504.17999},
  year={2025}
}

@article{wasi2024cogergllm,
  title={CogErgLLM: Exploring Large Language Model Systems Design Perspective Using Cognitive Ergonomics},
  author={Wasi, Azmine Toushik and Islam, Mst Rafia},
  journal={arXiv preprint arXiv:2407.02885},
  year={2024}
}

@article{guidroz2025llm,
  title={LLM-based Text Simplification and its Effect on User Comprehension and Cognitive Load},
  author={Guidroz, Theo and Ardila, Diego and Li, Jimmy and Mansour, Adam and Jhun, Paul and Gonzalez, Nina and Ji, Xiang and Sanchez, Mike and Kakarmath, Sujay and Bellaiche, Mathias MJ and others},
  journal={arXiv preprint arXiv:2505.01980},
  year={2025}
}

@article{cobbe2021gsm8k,
  title={Training Verifiers to Solve Math Word Problems},
  author={Cobbe, Karl and Kosaraju, Vineet and Bavarian, Mohammad and Chen, Mark and Jun, Heewoo and Kaiser, Lukasz and Plappert, Matthias and Tworek, Jerry and Hilton, Jacob and Nakano, Reiichiro and Hesse, Christopher and Schulman, John},
  journal={arXiv preprint arXiv:2110.14168},
  year={2021}
}

@article{hendrycks2021measuring,
  title={Measuring mathematical problem solving with the math dataset},
  author={Hendrycks, Dan and Burns, Collin and Kadavath, Saurav and Arora, Akul and Basart, Steven and Tang, Eric and Song, Dawn and Steinhardt, Jacob},
  journal={arXiv preprint arXiv:2103.03874},
  year={2021}
}

@misc{kojima2023large,
      title={Large Language Models are Zero-Shot Reasoners}, 
      author={Takeshi Kojima and Shixiang Shane Gu and Machel Reid and Yutaka Matsuo and Yusuke Iwasawa},
      year={2023},
      eprint={2205.11916},
      archivePrefix={arXiv},
      primaryClass={cs.CL},
      url={https://arxiv.org/abs/2205.11916}, 
}

@misc{zhang2022auto,
      title={Automatic Chain of Thought Prompting in Large Language Models}, 
      author={Zhuosheng Zhang and Aston Zhang and Mu Li and Alex Smola},
      year={2022},
      eprint={2210.03493},
      archivePrefix={arXiv},
      primaryClass={cs.CL},
      url={https://arxiv.org/abs/2210.03493}, 
}

@misc{gan2025rethink,
      title={Rethinking External Slow-Thinking: From Snowball Errors to Probability of Correct Reasoning}, 
      author={Zeyu Gan and Yun Liao and Yong Liu},
      year={2025},
      eprint={2501.15602},
      archivePrefix={arXiv},
      primaryClass={cs.AI},
      url={https://arxiv.org/abs/2501.15602}, 
}

@misc{hou2023mechanistic,
      title={Towards a Mechanistic Interpretation of Multi-Step Reasoning Capabilities of Language Models}, 
      author={Yifan Hou and Jiaoda Li and Yu Fei and Alessandro Stolfo and Wangchunshu Zhou and Guangtao Zeng and Antoine Bosselut and Mrinmaya Sachan},
      year={2023},
      eprint={2310.14491},
      archivePrefix={arXiv},
      primaryClass={cs.CL},
      url={https://arxiv.org/abs/2310.14491}, 
}

@misc{yang2018hotpot,
      title={HotpotQA: A Dataset for Diverse, Explainable Multi-hop Question Answering}, 
      author={Zhilin Yang and Peng Qi and Saizheng Zhang and Yoshua Bengio and William W. Cohen and Ruslan Salakhutdinov and Christopher D. Manning},
      year={2018},
      eprint={1809.09600},
      archivePrefix={arXiv},
      primaryClass={cs.CL},
      url={https://arxiv.org/abs/1809.09600}, 
}

@misc{trivedi2022musique,
      title={MuSiQue: Multihop Questions via Single-hop Question Composition}, 
      author={Harsh Trivedi and Niranjan Balasubramanian and Tushar Khot and Ashish Sabharwal},
      year={2022},
      eprint={2108.00573},
      archivePrefix={arXiv},
      primaryClass={cs.CL},
      url={https://arxiv.org/abs/2108.00573}, 
}

@misc{talmor2019common,
      title={CommonsenseQA: A Question Answering Challenge Targeting Commonsense Knowledge}, 
      author={Alon Talmor and Jonathan Herzig and Nicholas Lourie and Jonathan Berant},
      year={2019},
      eprint={1811.00937},
      archivePrefix={arXiv},
      primaryClass={cs.CL},
      url={https://arxiv.org/abs/1811.00937}, 
}

@misc{sap2019social,
      title={SocialIQA: Commonsense Reasoning about Social Interactions}, 
      author={Maarten Sap and Hannah Rashkin and Derek Chen and Ronan LeBras and Yejin Choi},
      year={2019},
      eprint={1904.09728},
      archivePrefix={arXiv},
      primaryClass={cs.CL},
      url={https://arxiv.org/abs/1904.09728}, 
}

@misc{gemma,
      title={Gemma 2: Improving Open Language Models at a Practical Size}, 
      author={Gemma Team and Morgane Riviere and Shreya Pathak and Pier Giuseppe Sessa and Cassidy Hardin and Surya Bhupatiraju and Léonard Hussenot and Thomas Mesnard and Bobak Shahriari and Alexandre Ramé and Johan Ferret and Peter Liu and Pouya Tafti and Abe Friesen and Michelle Casbon and Sabela Ramos and Ravin Kumar and Charline Le Lan and Sammy Jerome and Anton Tsitsulin and Nino Vieillard and Piotr Stanczyk and Sertan Girgin and Nikola Momchev and Matt Hoffman and Shantanu Thakoor and Jean-Bastien Grill and Behnam Neyshabur and Olivier Bachem and Alanna Walton and Aliaksei Severyn and Alicia Parrish and Aliya Ahmad and Allen Hutchison and Alvin Abdagic and Amanda Carl and Amy Shen and Andy Brock and Andy Coenen and Anthony Laforge and Antonia Paterson and Ben Bastian and Bilal Piot and Bo Wu and Brandon Royal and Charlie Chen and Chintu Kumar and Chris Perry and Chris Welty and Christopher A. Choquette-Choo and Danila Sinopalnikov and David Weinberger and Dimple Vijaykumar and Dominika Rogozińska and Dustin Herbison and Elisa Bandy and Emma Wang and Eric Noland and Erica Moreira and Evan Senter and Evgenii Eltyshev and Francesco Visin and Gabriel Rasskin and Gary Wei and Glenn Cameron and Gus Martins and Hadi Hashemi and Hanna Klimczak-Plucińska and Harleen Batra and Harsh Dhand and Ivan Nardini and Jacinda Mein and Jack Zhou and James Svensson and Jeff Stanway and Jetha Chan and Jin Peng Zhou and Joana Carrasqueira and Joana Iljazi and Jocelyn Becker and Joe Fernandez and Joost van Amersfoort and Josh Gordon and Josh Lipschultz and Josh Newlan and Ju-yeong Ji and Kareem Mohamed and Kartikeya Badola and Kat Black and Katie Millican and Keelin McDonell and Kelvin Nguyen and Kiranbir Sodhia and Kish Greene and Lars Lowe Sjoesund and Lauren Usui and Laurent Sifre and Lena Heuermann and Leticia Lago and Lilly McNealus and Livio Baldini Soares and Logan Kilpatrick and Lucas Dixon and Luciano Martins and Machel Reid and Manvinder Singh and Mark Iverson and Martin Görner and Mat Velloso and Mateo Wirth and Matt Davidow and Matt Miller and Matthew Rahtz and Matthew Watson and Meg Risdal and Mehran Kazemi and Michael Moynihan and Ming Zhang and Minsuk Kahng and Minwoo Park and Mofi Rahman and Mohit Khatwani and Natalie Dao and Nenshad Bardoliwalla and Nesh Devanathan and Neta Dumai and Nilay Chauhan and Oscar Wahltinez and Pankil Botarda and Parker Barnes and Paul Barham and Paul Michel and Pengchong Jin and Petko Georgiev and Phil Culliton and Pradeep Kuppala and Ramona Comanescu and Ramona Merhej and Reena Jana and Reza Ardeshir Rokni and Rishabh Agarwal and Ryan Mullins and Samaneh Saadat and Sara Mc Carthy and Sarah Cogan and Sarah Perrin and Sébastien M. R. Arnold and Sebastian Krause and Shengyang Dai and Shruti Garg and Shruti Sheth and Sue Ronstrom and Susan Chan and Timothy Jordan and Ting Yu and Tom Eccles and Tom Hennigan and Tomas Kocisky and Tulsee Doshi and Vihan Jain and Vikas Yadav and Vilobh Meshram and Vishal Dharmadhikari and Warren Barkley and Wei Wei and Wenming Ye and Woohyun Han and Woosuk Kwon and Xiang Xu and Zhe Shen and Zhitao Gong and Zichuan Wei and Victor Cotruta and Phoebe Kirk and Anand Rao and Minh Giang and Ludovic Peran and Tris Warkentin and Eli Collins and Joelle Barral and Zoubin Ghahramani and Raia Hadsell and D. Sculley and Jeanine Banks and Anca Dragan and Slav Petrov and Oriol Vinyals and Jeff Dean and Demis Hassabis and Koray Kavukcuoglu and Clement Farabet and Elena Buchatskaya and Sebastian Borgeaud and Noah Fiedel and Armand Joulin and Kathleen Kenealy and Robert Dadashi and Alek Andreev},
      year={2024},
      eprint={2408.00118},
      archivePrefix={arXiv},
      primaryClass={cs.CL},
      url={https://arxiv.org/abs/2408.00118}, 
}

@misc{liu2025spin,
      title={SpinQuant: LLM quantization with learned rotations}, 
      author={Zechun Liu and Changsheng Zhao and Igor Fedorov and Bilge Soran and Dhruv Choudhary and Raghuraman Krishnamoorthi and Vikas Chandra and Yuandong Tian and Tijmen Blankevoort},
      year={2025},
      eprint={2405.16406},
      archivePrefix={arXiv},
      primaryClass={cs.LG},
      url={https://arxiv.org/abs/2405.16406}, 
}

@misc{qwen2025qwen,
      title={Qwen2.5 Technical Report}, 
      author={Qwen and : and An Yang and Baosong Yang and Beichen Zhang and Binyuan Hui and Bo Zheng and Bowen Yu and Chengyuan Li and Dayiheng Liu and Fei Huang and Haoran Wei and Huan Lin and Jian Yang and Jianhong Tu and Jianwei Zhang and Jianxin Yang and Jiaxi Yang and Jingren Zhou and Junyang Lin and Kai Dang and Keming Lu and Keqin Bao and Kexin Yang and Le Yu and Mei Li and Mingfeng Xue and Pei Zhang and Qin Zhu and Rui Men and Runji Lin and Tianhao Li and Tianyi Tang and Tingyu Xia and Xingzhang Ren and Xuancheng Ren and Yang Fan and Yang Su and Yichang Zhang and Yu Wan and Yuqiong Liu and Zeyu Cui and Zhenru Zhang and Zihan Qiu},
      year={2025},
      eprint={2412.15115},
      archivePrefix={arXiv},
      primaryClass={cs.CL},
      url={https://arxiv.org/abs/2412.15115}, 
}

@inproceedings{wang2023does,
  title={Does physical adversarial example really matter to autonomous driving? towards system-level effect of adversarial object evasion attack},
  author={Wang, Ningfei and Luo, Yunpeng and Sato, Takami and Xu, Kaidi and Chen, Qi Alfred},
  booktitle={Proceedings of the IEEE/CVF international conference on computer vision},
  pages={4412--4423},
  year={2023}
}

@article{hao2024training,
  title={Training large language models to reason in a continuous latent space},
  author={Hao, Shibo and Sukhbaatar, Sainbayar and Su, DiJia and Li, Xian and Hu, Zhiting and Weston, Jason and Tian, Yuandong},
  journal={arXiv preprint arXiv:2412.06769},
  year={2024}
}

@article{chuang2024lookback,
  title={Lookback lens: Detecting and mitigating contextual hallucinations in large language models using only attention maps},
  author={Chuang, Yung-Sung and Qiu, Linlu and Hsieh, Cheng-Yu and Krishna, Ranjay and Kim, Yoon and Glass, James},
  journal={arXiv preprint arXiv:2407.07071},
  year={2024}
}

@article{zhao2024towards,
  title={Towards uncovering how large language model works: An explainability perspective},
  author={Zhao, Haiyan and Yang, Fan and Shen, Bo and Lakkaraju, Himabindu and Du, Mengnan},
  journal={arXiv preprint arXiv:2402.10688},
  year={2024}
}

@article{atanasova2023faithfulness,
  title={Faithfulness tests for natural language explanations},
  author={Atanasova, Pepa and Camburu, Oana-Maria and Lioma, Christina and Lukasiewicz, Thomas and Simonsen, Jakob Grue and Augenstein, Isabelle},
  journal={arXiv preprint arXiv:2305.18029},
  year={2023}
}

@article{liu2023towards,
  title={Towards understanding in-context learning with contrastive demonstrations and saliency maps},
  author={Liu, Fuxiao and Xu, Paiheng and Li, Zongxia and Feng, Yue and Song, Hyemi},
  journal={arXiv preprint arXiv:2307.05052},
  year={2023}
}

@article{wei2023larger,
  title={Larger language models do in-context learning differently},
  author={Wei, Jerry and Wei, Jason and Tay, Yi and Tran, Dustin and Webson, Albert and Lu, Yifeng and Chen, Xinyun and Liu, Hanxiao and Huang, Da and Zhou, Denny and others},
  journal={arXiv preprint arXiv:2303.03846},
  year={2023}
}

@article{wu2023analyzing,
  title={Analyzing chain-of-thought prompting in large language models via gradient-based feature attributions},
  author={Wu, Skyler and Shen, Eric Meng and Badrinath, Charumathi and Ma, Jiaqi and Lakkaraju, Himabindu},
  journal={arXiv preprint arXiv:2307.13339},
  year={2023}
}

@article{wei2022chain,
  title={Chain-of-thought prompting elicits reasoning in large language models},
  author={Wei, Jason and Wang, Xuezhi and Schuurmans, Dale and Bosma, Maarten and Xia, Fei and Chi, Ed and Le, Quoc V and Zhou, Denny and others},
  journal={Advances in neural information processing systems},
  volume={35},
  pages={24824--24837},
  year={2022}
}

@article{wu2025ctrls,
  title={CTRLS: Chain-of-Thought Reasoning via Latent State-Transition},
  author={Wu, Junda and Xiong, Yuxin and Li, Xintong and Hu, Zhengmian and Yu, Tong and Wang, Rui and Chen, Xiang and Shang, Jingbo and McAuley, Julian},
  journal={arXiv preprint arXiv:2507.08182},
  year={2025}
}

@article{wu2024ocean,
  title={OCEAN: Offline Chain-of-thought Evaluation and Alignment in Large Language Models},
  author={Wu, Junda and Li, Xintong and Wang, Ruoyu and Xia, Yu and Xiong, Yuxin and Wang, Jianing and Yu, Tong and Chen, Xiang and Kveton, Branislav and Yao, Lina and others},
  journal={arXiv preprint arXiv:2410.23703},
  year={2024}
}

@inproceedings{wu2024decot,
  title={Decot: Debiasing chain-of-thought for knowledge-intensive tasks in large language models via causal intervention},
  author={Wu, Junda and Yu, Tong and Chen, Xiang and Wang, Haoliang and Rossi, Ryan and Kim, Sungchul and Rao, Anup and McAuley, Julian},
  booktitle={Proceedings of the 62nd Annual Meeting of the Association for Computational Linguistics (Volume 1: Long Papers)},
  pages={14073--14087},
  year={2024}
}

@inproceedings{liu2024interact,
  title={Interact with the explanations: Causal debiased explainable recommendation system},
  author={Liu, Xu and Yu, Tong and Xie, Kaige and Wu, Junda and Li, Shuai},
  booktitle={Proceedings of the 17th ACM International Conference on Web Search and Data Mining},
  pages={472--481},
  year={2024}
}

@inproceedings{wu2022context,
  title={Context-aware information-theoretic causal de-biasing for interactive sequence labeling},
  author={Wu, Junda and Wang, Rui and Yu, Tong and Zhang, Ruiyi and Zhao, Handong and Li, Shuai and Henao, Ricardo and Nenkova, Ani},
  booktitle={Findings of the Association for Computational Linguistics: EMNLP 2022},
  pages={3436--3448},
  year={2022}
}

@inproceedings{wu2021deconfounded,
  title={Deconfounded and explainable interactive vision-language retrieval of complex scenes},
  author={Wu, Junda and Yu, Tong and Li, Shuai},
  booktitle={Proceedings of the 29th ACM International Conference on Multimedia},
  pages={2103--2111},
  year={2021}
}

\appendix

\newpage
\section{Model and Processing Details}
\label{sec:details}
We use publicly available instruction-tuned checkpoints: Gemma 2B Instruct (fast inference baseline), LLaMA 3.2B Instruct (balanced performance), and Qwen2.5 7B Instruct (strong reasoning quality). For each input, we prepend a standard reasoning prompt (e.g., ``Let's solve this step by step.'') and extract reasoning steps based on explicit markers (e.g., ``Step 1:''). Hidden states from the final transformer layer are projected to 128 dimensions before Gram matrix computation. Top-64 eigenvalues are retained per step to form spectral embeddings.

\subsection{Running Example: Llama-3B on SocialIQA}
To ground our abstraction and simulation framework, we walk through a single question from the SocialIQA benchmark as answered by Llama-3B. 

\textbf{Prompt}
\textit{
\{"id": "sample\_33", "prompt": "Context: Cameron flew on a plane because he figured he would get there faster than driving.Question: What will Cameron want to do next? Options:A) buy a ticket B) get back on the plane C) find a hotel Let's solve this problem step-by-step."\}}
\textbf{Model response}
\textit{\{"sample\_id": "sample\_33", "detected\_steps": 4, "steps": [\{"step\_id": 1, "text": "Step 1: Analyze the situation. Cameron flew on a plane because he thought it would be faster than driving. However, we don't know if he actually arrived faster or if he's now facing a new challenge."\}, \{"step\_id": 2, "text": "Step 2: Consider the options. The options are buying a ticket, getting back on the plane, or finding a hotel. Since Cameron is already on the plane, getting back on the plane is unlikely."\}, \{"step\_id": 3, "text": "Step 3: Evaluate the options. Buying a ticket is not necessary, as Cameron is already on the plane. Finding a hotel is a reasonable option, as it's common for people to stay overnight when they travel."\}, \{"step\_id": 4, "text": "Step 4: Choose the most logical option. Given the context, finding a hotel seems like the most logical next step. The final answer is: C) find a hotel."\}]\}}

The transition probabilities reveal that Llama-3B most frequently transitions from goal‐setting to hazard analysis before arriving at a conclusion, reflecting a two‐stage reasoning pattern. A t-SNE plot further confirms that these clusters are well‐separated in embedding space. This running example demonstrates how our abstraction captures both the content and dynamics of the model’s internal reasoning on a typical SocialIQA question.

\subsection{Simulating Reasoning Trajectories via Markov Rollouts}
% \noindent\textbf{Monte Carlo Simulation via Markov Transitions.} We simulate reasoning trajectories by sampling from the transition matrix \( P \), drawing each state from the conditional distribution:
% \[
% s_{t+1} \sim P(\cdot \mid s_t).
% \]

% This is equivalent to a Metropolis-Hastings algorithm~\cite{andrieu2003introduction} with identical proposal and target distributions, yielding acceptance probability:
% \[
% \alpha(s_t, s_{t+1}^*) = \min\left\{1, \frac{p_\theta(s_{t+1}^* \mid s_t) q(s_t \mid s_{t+1}^*)}{p_\theta(s_t \mid s_{t+1}^*) q(s_{t+1}^* \mid s_t)} \right\}.
% \]

% With \( p = q = P \), we have \( \alpha = 1 \), reducing the process to a standard Markov chain. The sampled trajectories \( \tau^{(i)} = (s_0^{(i)}, \dots, s_T^{(i)}) \) enable Monte Carlo estimation:
% \[
% \mathbb{E}_{\tau \sim P}[f(\tau)] \approx \frac{1}{N} \sum_{i=1}^N f(\tau^{(i)}),
% \]
% where \( f \) measures properties like state positions or transition frequencies. This reveals global patterns in latent reasoning.

To validate whether the learned transition model faithfully captures the temporal structure of reasoning, we simulate step-wise trajectories using the Markov matrix \( P \). Each trajectory is sampled by recursively drawing the next state from the conditional distribution:
\[
s_{t+1} \sim P(\cdot \mid s_t).
\]
This sampling process reduces to a standard Markov chain, where the acceptance probability is always 1 due to symmetric proposal and target distributions:
\[
\alpha(s_t, s_{t+1}^*) = \min\left\{1, \frac{p(s_{t+1}^* \mid s_t)}{p(s_t \mid s_{t+1}^*)} \right\} = 1.
\]

The resulting simulated trajectories \( \tau^{(i)} = (s_0^{(i)}, \dots, s_T^{(i)}) \) can be used to estimate trajectory-level statistics via Monte Carlo approximation:
\[
\mathbb{E}_{\tau \sim P}[f(\tau)] \approx \frac{1}{N} \sum_{i=1}^N f(\tau^{(i)}),
\]
where \( f \) may represent properties such as average step index or transition counts. These simulations allow us to probe whether the model captures coherent and temporally structured reasoning flows.
\begin{table*}[t]
  \centering
  \small
  
  \begin{tabular}{l l c c c c c r r}
    \toprule
    \textbf{Dataset} 
      & \textbf{Model} 
      & \textbf{C0}
      & \textbf{C1} 
      & \textbf{C2} 
      & \textbf{C3} 
      & \textbf{C4} 
      & \(\boldsymbol{\rho}\) 
      & \textbf{p-value} \\
    \midrule
            \multirow{3}{*}{SocialIQA}
      & Gemma & 2.26/1.99  & 3.20/2.74  & 1.46/1.56  & 3.56/6.35  & 2.76/2.27   
              & 1.000     & 0.001        \\
      & Llama & 3.59/3.09  & 1.99/1.58  & 3.88/6.47  & 1.16/1.16  & 2.98/2.14    
              & 1.000     & 0.001        \\
      & Qwen  & 4.43/3.06  & 1.38/1.59  & 3.65/2.31  & 2.54/1.95  & 4.67/6.41   
              & 1.000     & 0.001        \\
      \midrule            
    \multirow{3}{*}{CSQA}
      & Gemma & 1.73/3.06  & 3.05/6.47  & 2.85/2.68  & 2.54/3.74  & 1.67/2.11 
              & 0.700 & 0.188 \\

      & Llama & 2.08/1.65  & 3.62/3.06  & 3.11/2.07  & 3.95/6.46  & 1.16/1.18 
              & 1.000 & 0.001 \\

      & Qwen  & 1.29/1.31  & 4.35/3.20  & 2.41/1.81  & 4.48/6.49  & 3.63/2.27 
              & 1.000 & 0.001    \\
    \midrule
    \multirow{3}{*}{GSM8K}
      & Gemma & 3.02/3.03  & 2.69/3.98  & 1.77/2.16  & 2.68/3.38  & 3.02/6.56    
              & 0.700 &  0.188        \\
      & Llama & 2.44/2.06  & 1.19/1.17  & 3.10/3.15  & 1.69/1.56  & 3.12/6.50  
              & 1.000     & 0.001       \\
      & Qwen  & 4.30/3.51  & 1.47/1.47  & 3.59/2.65  & 4.52/6.51  & 2.43/2.04  
              & 1.000     & 0.001        \\
    \bottomrule
  \end{tabular}
\caption{Simulated vs. real average cluster positions and Spearman statistics.
Columns C0–C4 list mean simulated/real positions. Spearman $\rho$ measures the monotonic agreement between simulated and real rankings ($\rho = 1$ is perfect), and the p-value tests its significance ($p < 0.05$ indicates a non-random correlation).}
\label{tab:sim_real_full}
\end{table*}

\begin{figure}[htbp]
  \centering
  \includegraphics[width=\columnwidth]{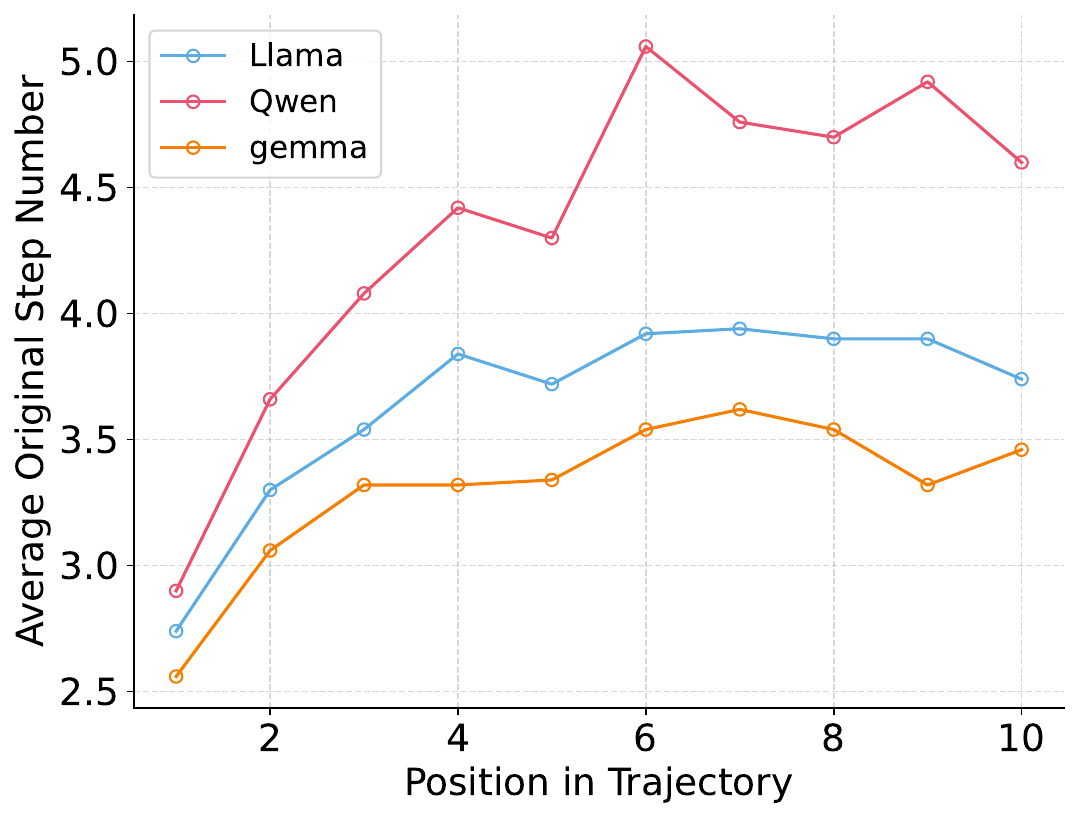}
  \caption{Average original step index at each position in the simulated trajectory on SocialIQA.} 
  \label{fig:stepindex}
\end{figure}

\subsection{Validating Temporal Consistency via Monte Carlo Simulation}

To evaluate whether the learned transition matrix captures the temporal structure of CoT reasoning, we perform Monte Carlo rollouts by sampling trajectories from the Markov model. Starting from a common initial cluster, we generate 10-step latent sequences based on \( P \), and analyze the correspondence between simulated states and their real positions in CoT.

Figure~\ref{fig:stepindex} shows the average original step index of reasoning steps sampled at each simulated position. The upward trend across all models indicates that our transition model preserves the directional nature of reasoning—from early-stage states (e.g., problem framing) to later stages (e.g., synthesis). This validates the ability of \( P \) to approximate real-world temporal dynamics in CoT trajectories.

Beyond position-wise trends, we also examine the expected position of each cluster across simulated trajectories. We estimate \( \mathbb{E}_{\tau \sim P}[f(\tau)] \) to obtain the average step index for each cluster and compare it against empirical values. Table~\ref{tab:sim_real_full} reports these results, along with Spearman correlation scores. For all models and datasets, we observe near-perfect alignment between simulated and real rankings, confirming that the learned abstraction preserves both semantic and temporal consistency.

\end{document}